\pdfoutput=1

\documentclass[11pt]{article}

\usepackage[]{EMNLP2023}

\usepackage{times}
\usepackage{latexsym}

\usepackage[T1]{fontenc}

\usepackage[utf8]{inputenc}

\usepackage{microtype}

\usepackage{inconsolata}

\usepackage{graphicx}
\usepackage{colortbl}
\usepackage{xcolor,soul}	%
\usepackage{algorithm}
\usepackage[noend]{algpseudocode}
\usepackage{xspace}
\usepackage{amsmath}
\usepackage{multirow}

\newcommand{\eat}[1]{}

\newcommand{\dmax}{d_{\mathrm{max}}}

\newcommand{\negs}{\mathit{negs}}

\newcommand{\gapxx}{\hspace*{4mm}}

\newcommand{\gapxxxx}{\hspace*{8mm}}

\newcommand{\sysname}{\textsc{Reflex}\xspace}
\newcommand{\bfit}[1]{\textbf{\textit{#1}}}

\usepackage{quoting}
\usepackage{amssymb}  %
\newenvironment{myquote}{                   %
  \parskip 0mm \begin{quoting}[vskip=0mm,leftmargin=2mm]}{
\end{quoting}}
\newenvironment{ite}{                     %
     \parskip 0cm \begin{itemize} \parskip 0cm \parsep 0cm \itemsep 0cm \topsep 0cm}{
        \end{itemize}} %
\newenvironment{enu}{                   %
     \parskip 0cm \begin{list}{}{\parsep 0cm \itemsep 0cm \topsep 0cm}}{
       \end{list}} %
\newenvironment{des}{                 %
     \parskip 0cm \begin{list}{}{\parsep 0cm \itemsep 0cm \topsep 0cm}}{
       \end{list}} %

\title{Language Models with Rationality} %

\newcommand{\authorsep}{\hspace{2ex}}

\author{
  Nora Kassner\textsuperscript{1,2} \authorsep 
  Oyvind Tafjord\textsuperscript{1} \authorsep 
  Ashish Sabharwal\textsuperscript{1} \authorsep 
  Kyle Richardson\textsuperscript{1} \\ %
  {\bf Hinrich Sch{\"u}tze\textsuperscript{2} \authorsep 
  Peter Clark\textsuperscript{1}} \\ %
  \hspace{1ex} \\
\textsuperscript{1}Allen Institute for AI, Seattle, WA \\
\textsuperscript{2}Center for Information and Language Processing, 
LMU Munich, Germany \\
\texttt{kassner@cis.lmu.de}\\
\texttt{\{oyvindt,ashishs,kyler,peterc\}@allenai.org} 
}

\begin{document}
\maketitle

\begin{abstract}
While large language models (LLMs) are proficient at question-answering (QA), it is
not always clear how (or even if) an answer follows from their latent ``beliefs''.
This lack of interpretability is a growing impediment to widespread use of LLMs.
To address this, our goals are to make model beliefs and their inferential
relationships explicit, and to resolve inconsistencies that may exist,
so that answers are supported by interpretable chains of reasoning
drawn from a consistent network of beliefs.
Our approach, which we call \sysname, is to add a
{\bf rational, self-reflecting layer} on top of the LLM.
First, given a question, we construct a {\bf belief graph} using a backward-chaining process to
materialize relevant model beliefs (including beliefs about answer candidates) and their inferential relationships.
Second, we identify and minimize contradictions in that graph using a formal constraint reasoner.
We find that \sysname significantly improves consistency (by 8\%-11\% absolute) without harming overall answer accuracy,
resulting in answers supported by faithful chains of reasoning drawn from a more consistent belief system.
This suggests a new style of system architecture in which an LLM extended
with a rational layer can provide an interpretable window into system beliefs,
add a systematic reasoning capability, and repair latent inconsistencies present
in the LLM.
\end{abstract}

\begin{figure}[t]
\centering
     \includegraphics[width=1\columnwidth]{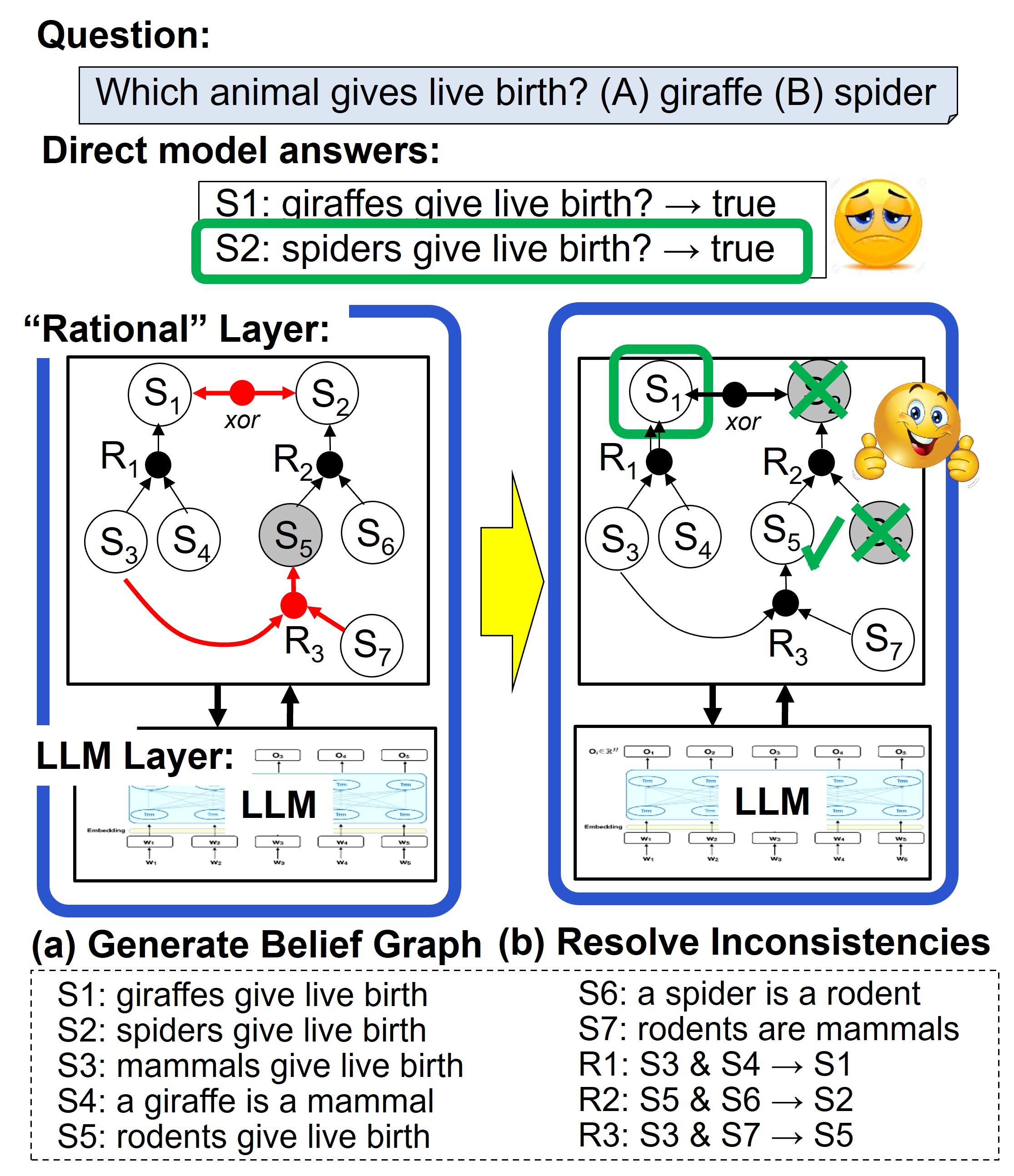}	   %
       \vspace{-7mm}
\caption{
({\bf Top}) When queried about each answer option independently, the model incorrectly believes both are true, and is more confident in the wrong answer ($S_{2}$).
({\bf Bottom}) \sysname{} adds a "rational" layer above the LLM layer, in which a {\bf belief graph} is constructed
(by iteratively querying the LLM, up/down arrows),
containing relevant model-believed facts (white/grey = believed T/F) and their inferential relationships. Inconsistencies are then identified (red) and minimized by a constraint reasoner that flips T/F labels on beliefs (green $\checkmark$/X), here resulting in the correct answer ($S_{1}$, green box)
+ explanation (graph) by the overall system (blue).
\label{example}}
\vspace{-4mm}

\end{figure}

\vspace*{-1mm}
\section{Introduction}
\begin{figure*}[ht!]
\centering
     \includegraphics[width=0.9\textwidth]{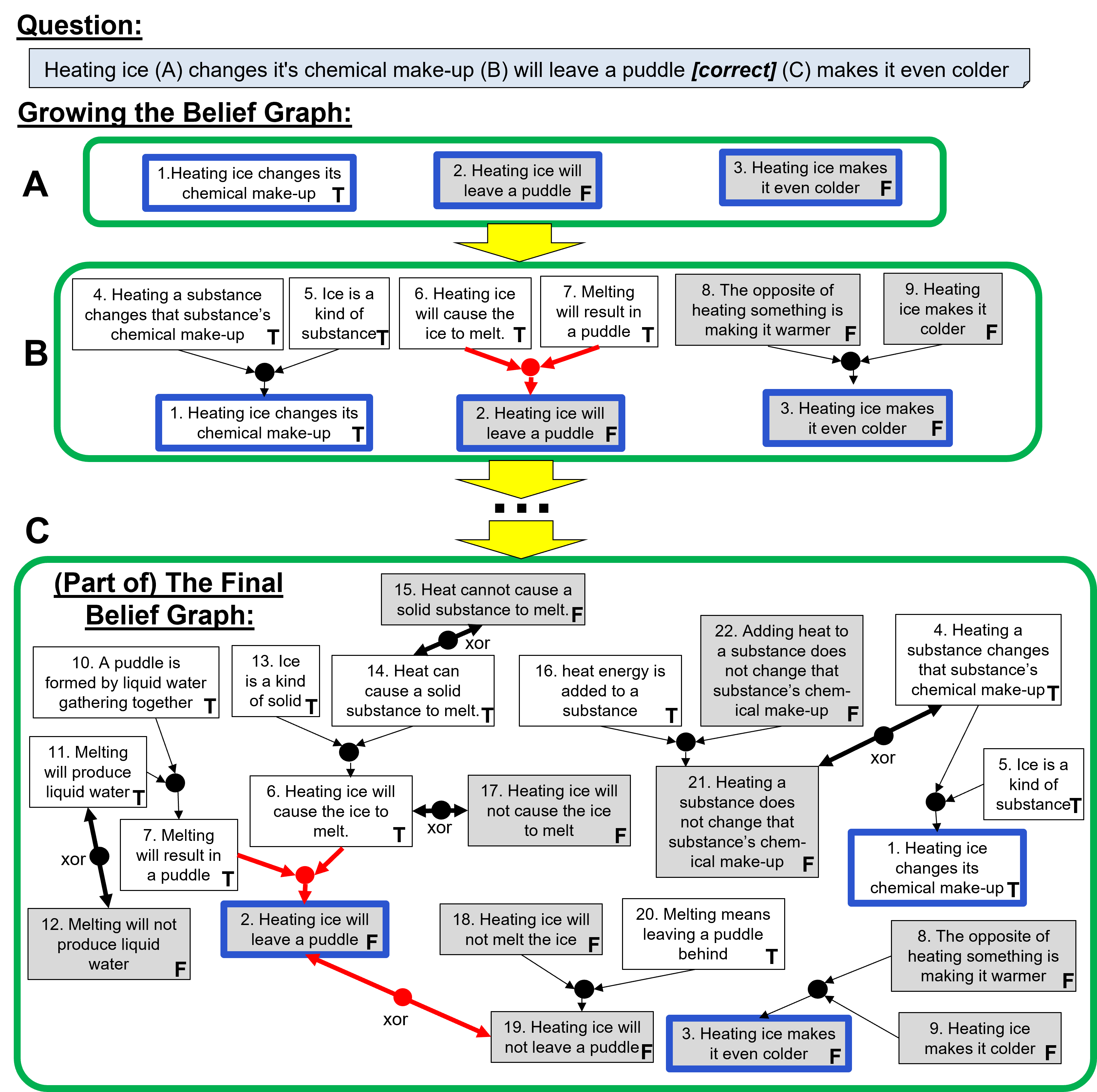}	   %
\caption{
Given a question, each answer choice is first converted to a hypothesis statement ({\bf A}).
The belief graph is then constructed in stages, first generating rules that conclude the hypotheses ({\bf B}),
then backward-chaining to generate rules concluding the premises of those first rules, etc., 
and adding in negated versions of graph statements connected with the originals via XOR links (e.g., nodes 11 and 12),
until the stopping criterion is met ({\bf C}). 
Statements are then labeled with the model's belief in them (true/false),
found via self-querying (white = believed true, grey = believed false). Finally, logical conflicts are identified
(colored red), and constraint satisfaction techniques are used to resolve them.
In this case, as there is strong evidence that node 2 is actually true (7 \& 6 $\rightarrow$ 2, not(19) $\rightarrow$ 2),
the solver finds that the minimum cost repair is to flip node 2's label from FALSE to TRUE. Here, node 2 ends up being
selected as the final answer, thus correctly answering the original question.
\label{graph-growth}}
\end{figure*}

\vspace*{-1mm}
While large language models (LLMs) are impressive at question-answering (QA), it is
not always clear how (or even if) an answer follows from their latent ``beliefs''\footnote{\label{footnote:belief}
	   We adopt a simple definition of belief, namely that a model
	   believes X if it answers "yes" to the question "Is X true?".
	   Other definitions could also be used; see Section~\ref{related-work}.}
about the world, or whether the LLM even has a coherent internal belief system.
This general opacity is a growing impediment to widespread use of LLMs, e.g.,
in critical applications such as medicine, law, and hiring decisions,
where properties of explainability, interpretability, and trust are
paramount. Our goal is to help alleviate such opacity by
constructing an explicit representation of 
system beliefs and their inferential relationships (including to answer candidates),
so that answers are supported by interpretable chains of reasoning.
These constructed {\bf belief graphs}, e.g., Figures~\ref{example} and~\ref{graph-growth},
form a {\bf rational layer} above the LLM explaining how answers follow from beliefs,
and provide a window into some of the latent contents of the model, potentially
helping users understand and trust model answers.

In addition, when we do this, we find such graphs expose latent inconsistencies
in the model's beliefs. We show how such inconsistencies can be resolved 
using constraint satisfaction techniques.
When we do this, the rational layer becomes not just a window onto
the model, but an active reasoning component in its own right in a {\it larger, overall
system}, comprising the (frozen) LLM plus rational layer
 (blue box, Figure~\ref{example}). We show 
this results in a more consistent set of beliefs in the overall system,
without harming overall answer accuracy (although some individual answers
may change). The result is answers supported by faithful, system-believed
chains of reasoning drawn from a consistent belief system. %

Our approach, called \sysname, introduces a \textbf{rational layer} consisting of two parts.
First, to produce a \underline{belief graph}, we recursively ask the LLM to explain
why each candidate answer
might be true, expressed as a set of sentences that entail the answer.
This builds on earlier work on generating entailment-based and
chain-of-thought explanations \cite{entailer,nellie,Wei2022ChainOT}.
We then add a self-verification step to check that the model itself
believes those generations (i.e., that the model believes what it says),
allowing us to identify sentences reflecting the model's own internal
knowledge. For example, in Figure~\ref{example}, when asked to
explain S1 (``giraffes give live birth''), the model generates S7
([because] ``mammals give live birth'') and S4 ([and] ``a giraffe is a mammal'').
Self-querying then checks if the model actually believes its generations
(``Do mammals give live birth?''). The answer (``yes''/''no'') assigns a
true/false (T/F) value to each generation,
indicated in Figure~\ref{example} by white/grey nodes. This procedure is then
applied recursively to the generated, supporting sentences.
The resulting
network of model beliefs and their dependencies provides a
a window into the model.

Second, we apply a formal \underline{constraint reasoner} to this graph to
resolve inconsistencies, by finding the optimal (minimal cost,
Section~\ref{reasoning}) way of flipping T/F values.
For example, on the left in Figure~\ref{example}, S2 and S3 (``spiders
do/don't give live birth'') are in an XOR relationship (i.e., exactly one must be false),
but both are believed as true (white) by the LLM -
a latent contradiction within the LLM. Constraint reasoning
then seeks to remove such inconsistencies, here 
flipping the belief value on S2 from T to F (Figure~\ref{example}, right),
repairing the contradiction. This builds on earlier techniques
\cite{Kassner2021BeliefBankAM,Mitchell2022EnhancingSA,jung-etal-2022-maieutic},
though in a notably richer setting
with over 350 nodes and 80 constraints per question, joint inference across answer candidates,
and a variety of constraint types.
The overall result is a fully autonomous, self-reflective
system that is able to deliberate (and if necessary change) its answers,
thereby resolving latent inconsistencies that would otherwise go unnoticed, and
provide %
faithful explanations drawn from a consistent belief system.

We evaluate our implementation of \sysname on three datasets:
EntailmentBank \cite{Dalvi2021ExplainingAW}, OBQA \cite{obqa}, and QuaRTz \cite{quartz}.
We find that \sysname is able to construct belief graphs with significantly
improved consistency (by 8\%-11\% absolute) without harming overall answer accuracy.
In addition, answers are now supported by
a more consistent, system-believed chain of reasoning, providing 
a window into the previously latent beliefs of the model.
Our contributions are thus:
\begin{enu}
\item[1.] A {\bf new style of system architecture} in which an LLM is
  extended with a {\bf rational layer} in which an explicit representation
  of system beliefs and relationships is constructed and which can
  be reasoned over. This layer provides an {\bf interpretable window} into
  system beliefs, adds a systematic reasoning capablity, and allows latent
  inconsistencies present in the LLM to be repaired.
\item[2.] An implementation of this architecture demonstrating that the
  {\bf consistency of the overall system's network of beliefs can be
  significantly improved} without harming answer accuracy. Answers
  are now supported by explicit, interpretable chains of reasoning drawn from a more
  consistent network of beliefs. %
\end{enu}

\vspace*{-1mm}
\section{Related Work \label{related-work}}
\vspace*{-1mm}
{\bf Materializing a Model's Internal Knowledge:} 
It is now well recognized that LLMs contain extensive world knowledge
\cite{lama,petroni2020how,davison-etal-2019-commonsense,peters2019knowledge,alpaqa,roberts2020much}
that somehow enables them to perform well.
Recent work has attempted to expose that knowledge in various ways,
both to justify answers and improve performance, and our work falls into this genre.
Standard explanation generation methods \cite{Wiegreffe2021TeachMT} can produce compelling
explanations, but with no guarantee that the generated sequence of tokens
expresses the model's internal knowledge, nor entails the actual answer.
Similarly, chain-of-thought (CoT) \cite{Wei2022ChainOT} and Least-to-Most \cite{Zhou2022LeasttoMostPE}
prompting generate (in different ways) a step-by-step reasoning chain along with
an answer, but again with no claim that the chain
reflects the model's internal knowledge nor is valid reasoning
\cite{Subramanian2020ObtainingFI}.

To add semantics to generations, several systems have used self-querying
to verify that generations reflect model-believed facts (by self-querying ``Is p true?'') \citep[e.g.,][]{Kassner2021BeliefBankAM,jung-etal-2022-maieutic}, or
model-believed rules (by self-querying ``Does p imply q?'') \citep[e.g.,][]{entailer}.
We build on these to construct a {\bf belief graph},
namely a network of model-believed facts and their inferential relationships,
which can then be reflected on.

{\bf Beliefs:} We refer to the model's factual opinions as ``beliefs'' rather
than ``knowledge'' because those opinions may be wrong. In general,
an agent can be said to believe p if it acts as if p was true \cite{sep-belief}.
Following \citet{Kassner2021BeliefBankAM} and \citet{richardson2022breakpoint},
we take a simple, syntactic operationalization  of this, namely the agent answers ``yes'' to the question ``p?'',
but also note that more semantic versions could be used, e.g.,
the agent also answers ``yes'' to paraphrases and implications of p.

{\bf Reducing Inconsistency:} LLMs are known to be inconsistent in their answers
\cite{ettinger-2020-bert, Kassner2020NegatedAM, davison-etal-2019-commonsense,ravichander-etal-2020-systematicity,
  Elazar2021MeasuringAI,Subramanian2020ObtainingFI,Gu2022DoLM},
and several recent works have used constraint reasoners to identify and reduce inconsistency.
BeliefBank used a MaxSAT solver to resolve inconsistencies between model beliefs, but
required a hand-provided set of constraint rules \cite{Kassner2021BeliefBankAM}.
ConCoRD \cite{Mitchell2022EnhancingSA} similarly used MaxSAT 
to ensure model answers were consistent with NLI-derived entailment
constraints between them, but did not introduce additional model-believed
facts and rules. Maieutic Prompting \cite{jung-etal-2022-maieutic} also used
MaxSAT to resolve inconsistencies between facts in prompt-induced explanation chains. However, those chains were not validated as reflecting 
model-believed constraint rules\footnote{\sysname checks whether both the statements $s_i$, and the rules ($s_i \rightarrow h$), are believed by the model via self-querying, e.g., by asking “Does $s_i \rightarrow h$?”, and also scores the strength of those beliefs. In maieutic prompting, the generated rules are not checked against the model, resulting in rules that the model itself may not believe, if queried about them.},
and did not support conjunction.
\sysname extends these reasoning chains to provide a
full semantic account of how answers are supported by the model's
internal knowledge. Additionally, it performs joint reasoning across answer candidates
and operates at a much larger scale
(e.g., over 350 nodes on average for each question) and with a variety of constraint types.

\section{\sysname: Our Approach}

\subsection{Belief Graphs}
\label{subsec:belief-graphs}

Our belief graphs are defined over a set of natural language true/false \emph{statements} and represent a set of \emph{rules} that constrain the truth values of these statements. We refer to statements that are factually true in the world as \emph{facts}. The truth value assigned by a model $M$ to a statement is referred to as $M$'s \emph{belief} in that statement (cf.~Footnote~\ref{footnote:belief}). A model's internal beliefs may not always align with facts. Our goal is to extract a model's initial beliefs about statements inferentially related to all top-level hypotheses of interest, and perform reasoning to update these beliefs so as to make them more consistent with respect to the rules, and ideally also factually more accurate.

A belief graph is a type of \emph{factor graph} commonly used in the probabilistic inference literature~\cite{Loeliger2004AnIT}. Formally, it is defined as an undirected graph $G = (N,E)$ with nodes $N$ and edges $E$.
Nodes are of two types: A {\it statement node} (referred to as a "variable node" in a factor graph) is a triple $(s,l,c_{s})$
containing a natural language statement $s$, an associated value $l \in \{T,F\}$ initially
denoting $M$'s belief that $s$ is true or false, and a confidence $c_{s} \in [0,1]$ denoting a confidence in that label.
A {\it rule node} (referred to as a "factor node" in a factor graph) is a pair $(r,c_r)$ denoting a disjunctive rule or constraint over statements, with confidence $c_r$.
It takes the form $r = (-s_1 \lor \ldots \lor -s_\ell \lor s_{\ell+1} \lor \ldots \lor s_k)$. For ease of interpretation, we view this constraint as $r = p \rightarrow h$ where $p = s_1 \land \ldots \land s_\ell$ is a conjunctive premise and $h = s_{\ell+1} \lor \ldots \lor s_k$ is a disjunctive hypothesis. The rule says that if $p$ is true, so must be $h$; and the contrapositive of this.

Edges $E$ connect rule nodes to the statements they constrain, denoting their dependence.
For legibility, we draw edges directionally to depict the way the rule reads: the statements in $p$ point to $r$, which in turn points to $h$.
Mathematically, the influence is bidirectional and the depicted directionality is irrelevant during reasoning (Section~\ref{reasoning}), just as in a standard factor graph.

\newcommand{\cost}{\mathit{cost}}

We adopt the standard probabilistic semantics of factor graphs, thereby associating a belief graph with a well-defined probability distribution over any set of statement beliefs. For a \textbf{statement node} $(s,l,c_s)$, the cost $\cost_s$ for setting it to $l$ is $0$, and that for setting it against $l$ is $c_s$; the corresponding \emph{weight} of this node is $w_s = \exp(-\cost_s)$. Costs and weights for a \textbf{rule node} $(r,c_r)$ are defined similarly, based on whether the beliefs satisfy $r$ or not. Finally, the overall weight of a T/F assignment to all statements is $\prod_s w_s \cdot \prod_r w_r$, which, when normalized by the total weight across all possible assignments, yields a probability distribution over such assignments. We will be interested in finding the \emph{most consistent set of beliefs}, i.e., a T/F assignment to statements with the minimum overall weight, which is equivalent to minimizing $\sum_s \cost_s + \sum_r \cost_r$. This is referred to as the MPE (most probable explanation) problem in the graphical models literature, which we later solve exactly using a MaxSAT constraint solver based on a standard translation of MPE into weighted MaxSAT \cite{park2002using,sang2007dynamic}.

\subsection{Constructing Belief Graphs}

Given an initial node (statement) $s$, a belief graph $G$ is produced by a backward-chaining process described below,
in which $G$ is recursively expanded to add statements that together may entail $s$.

\subsubsection{Basic Operations}
\label{subsec:basic-ops}

\newcommand{\score}{\mathrm{score}}

Let $h$ denote a hypothesis (language statement $s$) of interest and $p$ a premise---a set of statements \{$s_1$,\ldots,$s_n$\} that together \emph{may} entail $h$.
Given these, there are {\bf three basic operations} required to generate belief graphs:
\begin{des}
\item[\textbf{1.} $h \Rightarrow p$:] Given $h$, {\it generate} a $p$ that may entail $h$.

\item[\textbf{2.} $s \Rightarrow (l, c_s)$:] Given a statement $s$, output a true/false value $l$ and a confidence in the belief that $s$ has truth value $l$ (as assessed via yes/no question-answering). 

\item[\textbf{3.} ${(p,h) \Rightarrow c_{r}}$:] Given $p$ and $h$, output a confidence that the candidate rule $r =  p \rightarrow h$ holds.
\end{des}

The most important of these is the first operation, in which the model self-generates conjunctive
rules concluding $h$ (i.e., reason $p$ for believing $h$),
thus adding new nodes to the graph.

There are several ways of implementing these basic functions,
and our algorithm is agnostic to the method used. In our work here,
we use Entailer, an off-the-shelf T5-11B trained model with these functionalities \cite{entailer}. Further, since the raw score produced by the model tends to be skewed towards 0 or 1, when computing $c_s$ and $c_r$ in practice, we re-scale the raw model score using a set of hyperparameters (cf.~Appendix~\ref{appendix:hyperparams-and-runtime}).

One may use alternative ways to implement these operators, such as chain-of-thought prompting a model like GPT3 \cite{Wei2022ChainOT} or ChatGPT \cite{chatgpt}.
For example, to generate a rule concluding a hypothesis $h$ such as ``Plants require CO2 to make their food.'', the model
could be prompted with $h$ followed by ``Explain the last statement with a 2-step reasoning chain.'',
the numbered generations forming the premise $p$.
Similarly, generated statements and rules
can be validated as reflecting the model's beliefs by self-querying
(``Is $s$ true?'', ``Does $p$ imply $h$?''), and then using the generated yes/no answer token
probabilities as the model's confidence \cite{Kadavath2022LanguageM}.

\subsubsection{Initial Hypothesis Generation \label{hypothesis-generation}}

Given a question, we first generate a set $\mathcal{H}$ of hypothesis sentences
(e.g., ``Is the sky (A) blue (B) yellow'' $\rightarrow$
\{ $h_1$ = ``The sky is blue.'', $h_2$ = ``The sky is yellow.'').\footnote{Conversion of a QA pair to a declarative hypothesis D uses a custom T5-11B model trained on the QA2D dataset \cite{qa2d}.}
An $N$-way multiple choice question yields $N$ hypotheses in $\mathcal{H}$.
A true/false question yields 2 hypotheses. To handle open-ended questions,
candidate answers can be generated, e.g., using 
nucleus sampling \cite{nucleus-sampling}.

\begin{algorithm}[t]
\begin{algorithmic}[1]
\small{
\Procedure{generate-graph}{hypotheses $\mathcal{H}$, max depth $\dmax$}:
\State let $G =$ empty graph
\State {\bf foreach} $h \in \mathcal{H}$
\State \gapxx {\bf call} \textsc{extend-graph}($h, 0, \dmax$, $G$)
\State {\bf add} MC rule node $\left( \bigvee_{h \in \mathcal{H}} h, \infty \right)$ to $G$
\State {\bf foreach} pair $(h_i, h_j)$ of hypotheses in $\mathcal{H}$
\State \gapxx {\bf add} MC rule node $(\neg h_i \vee \neg h_j, c_{\mathrm{mc}})$ to $G$
\State \Return $G$
\EndProcedure

\item[]   %

\Procedure{extend-graph}{statement $s$, current depth $d$, max depth $\dmax$, partial graph $G$}:
\State {\bf call} operator $s \Rightarrow (l, c_s)$ to score statement $s$
\State {\bf add} statement node $(s, l, c_s)$ to $G$
\State {\bf gen.} the negation sentence $\negs$ = neg($s$)
\State {\bf add} rule node $(\mathrm{XOR}(s,\negs), c_{\mathrm{xor}})$ to $G$
\State {\bf call} \textsc{extend-graph}($\negs, d+1, \dmax$, $G$)
\State {\bf if} $d < \dmax$ {\bf do}:
\State \gapxx let $h = s$
\State \gapxx {\bf call} operator $h \Rightarrow p$ to generate $p$
\State \gapxx {\bf call} operator $(p,h) \Rightarrow c_r$ to score rule $p \to h$
\State \gapxx {\bf add} rule node $(p \to h, c_r)$ to $G$
\State \gapxx {\bf foreach} $s_i \in p$
\State \gapxxxx {\bf call} \textsc{extend-graph}($s_i, d+1, \dmax$, $G$)
\EndProcedure
}
\end{algorithmic}
\caption{\label{bg-algorithm} The recursive algorithm for constructing a belief graph of max depth $\dmax$ for a hypothesis set $\mathcal{H}$. The subroutine \textsc{extend-graph} takes a partial graph $G$ as an input and extends it in place with one statement and its subgraph.
}
\end{algorithm}

\subsubsection{Belief Graph Generation \label{belief-graph-generation}}

The belief graph generation process is shown in Algorithm~\ref{bg-algorithm}. 
An example of (part of) a generated belief graph is shown in Figure~\ref{graph-growth}.

Given a set $\mathcal{H}$ of hypotheses, we generate a single belief graph $G$
by using our basic operations (Section~\ref{subsec:basic-ops}) to recursively generate rules that conclude each $h_{i} \in \mathcal{H}$ up to a fixed maximum depth $\dmax$. 
(Each original $h_i$ is at depth $d = 0$.)

For each statement $s$, we also generate nodes $\negs$ (and their recursive subgraphs) expressing its negation, e.g., ``The sky is not blue.'' from ``The sky is blue.''.\footnote{
  We use a simple, custom-built utility for this, namely a T5-base model trained on 9k Turk-generated examples.}
Each pair $s$ and $\negs$ is connected with an XOR rule, indicating a (soft) preference for setting exactly one of them to true;
this is represented as two disjunctive constraints $(s \lor \negs)$ and $(-s \lor -\negs)$ whose weight $c_{\mathrm{xor}}$ is a fixed hyperparameter.
Lastly, we add a multiple-choice (MC) constraint which has two parts: a hard constraint (with infinite cost)
  that at least one hypothesis must be chosen, and a soft constraint\footnote{soft, to allow for cases with multiple valid answers, e.g., open-ended questions or those asking for the best answer.}
  that no more than one should be chosen. The soft constraint is associated with a fixed hyperparameter weight $c_{\mathrm{mc}}$.

\subsection{Reasoning Over Belief Graphs \label{reasoning}}

Belief graphs provide a window into the model's beliefs about some of the relevant statements
and their (believed) inferential relationships to candidate answers to a question.
As others have shown \cite{Kassner2021BeliefBankAM,Mitchell2022EnhancingSA}, such beliefs can be inconsistent, and materializing those inconsistencies provides one the opportunity to remove or reduce them.

In a similar vein, and as discussed in Section~\ref{subsec:belief-graphs}, \sysname performs inference over belief graphs in order to compute an updated set of beliefs that is as consistent as possible with the rules. To this end, it converts belief graphs into an equivalent weighted MaxSAT problem and uses an off-the-shelf MaxSAT solver (RC2, \cite{Ignatiev2019RC2AE})
to compute the optimal flips of initial
true/false beliefs that minimize global inconsistency.
It then discards all rules that are in conflict with the updated statement beliefs, obtaining a smaller, updated belief graph.
This \textbf{smaller belief graph produced by \sysname is self-consistent} and provides
inferential support for the top-level hypotheses.

\subsection{Generating Faithful Explanations}

Notably, the smaller updated belief graph produced by \sysname provides a \textbf{faithful} explanation of the answer it predicts, in the sense that it accurately represents the reasoning process behind {\it the overall system's} prediction~\cite{lyu2022faithful}.
This is true as the MaxSAT reasoning process results precisely in a self-consistent set of beliefs from which \sysname determines whether to believe a candidate answer or not, and produces its final prediction based on this (rather than on the raw LLM output alone; 
note that we do not make any claims about how the internal reasoning of the LLM component operates.)
Thus, \sysname provides the user with an
interpretable reasoning trace, allowing the user to understand how it derived the answer
from more rudimentary facts~\cite{Subramanian2020ObtainingFI}. %

We note that the original belief graph (before reasoning) may reveal that the model's
original explanation is, in fact, \emph{not} faithful to its own beliefs.
For example, in Figure~\ref{graph-growth},
the model believes statements 6, 7, and that 6 \& 7 entail 2, but does not
believe 2 (colored grey).
Thus, the global reasoning layer of \sysname plays a critical role in arriving at faithful explanations.

\section{Experiments and Results}
\label{sec:experiments}

The goal of our experiments is to evaluate the extent to which our overall system, namely an LLM plus a self-reflecting, rational layer, helps expose and resolve inconsistencies 
in the LLM's beliefs without harming accuracy.
Importantly, \sysname is evaluated in a \emph{zero-shot} setting, without relying on training instances of the target datasets.

\paragraph{Datasets.}
We use the test partitions of three existing multiple-choice datasets:
EntailmentBank \cite{Dalvi2021ExplainingAW}, OBQA \cite{obqa}, and QuaRTz \cite{quartz}. We chose our datasets as they contain inferentially rich questions (typically) requiring reasoning.
The partitions contain
339, 500, and 784 examples, respectively.

\paragraph{Models.}
The {\bf baseline LLM} we use is an LLM that has been trained to perform QA and also supports the basic operations discussed in Sec.~\ref{subsec:basic-ops}, enabling us to assess how much it can be improved by adding a \sysname layer. To this end, we use
a publicly available, frozen, off-the-shelf T5-11B LLM called Entailer \cite{entailer}. To answer an MC question with this LLM, we score each answer hypothesis ($c_{s}$, Section~\ref{subsec:basic-ops}) and select the one with the highest truth confidence. If Entailer assigns false values to all answer choices, we select the hypothesis with the lowest false confidence.

{\bf \sysname} then adds a rational layer to this LLM, creating a new system that is also able to self-reflect and modify its beliefs.
To ensure the different belief graph scores in \sysname are appropriately calibrated, we
use nine hyperparameters, tuned once on the dev partition of
EntailmentBank \cite{Dalvi2021ExplainingAW} and then
kept fixed for all experiments. Details are in Appendix~\ref{appendix:hyperparams-and-runtime}.
Note the LLM itself remains frozen, with 
belief revision occurring in the rational (belief graph)
layer above it.

\paragraph{Metrics.}
For measuring \textbf{self-consistency}, we follow \citet{Li2019ALF} and report the \emph{conditional 
constraint violation} ($\tau$) metric, defined as follows:
the fraction of rules whose {\it premises} $p$ are believed true,
but whose {\it hypothesis} $h$ is not. In other words, over all rules of the form $p \to h$, $\tau$ is:
\begin{align*}
    \tau = \frac{| \{ p \to h \mid p = \mathrm{T}, h = \mathrm{F} \} |} {| \{ p \to h \mid p = \mathrm{T} \} |}
\end{align*}
where $s = T$ denotes the system believes statement $s$ to be true (similarly for $s = F$). The numerator of $\tau$ thus captures the number of constraints the system \emph{violates}. The denominator captures the number of \emph{applicable} constraints.
We then report the following metric: \textbf{consistency} = 1 - $\tau$.

For {\bf QA performance}, we report standard \textbf{multiple-choice accuracy}: 1 point for predicting the correct answer, 1/$N$ points for predicting $N$ answers including the correct one, 1/$k$ points for no prediction ($k$ = \# answer options), 0 otherwise.

\subsection{Results}
\paragraph{Consistency.} Table \ref{Results-overall} shows consistency results on the test partitions of our datasets.
We observe {\bf significant consistency gains} (by 8\%-11\% absolute), showing \sysname's effectiveness at creating a consistent belief network within the overall system.

\begin{table}[ht]
\centering
\setlength{\tabcolsep}{2pt}	%
\begin{tabular}{l|c|c|c} \hline
   & Entail- &  &  \\
System & mentBank & OBQA & Quartz \\ \hline  
LLM & 87.0 & 88.2 & 85.7\\ 
LLM + rational layer & \multirow{2}{*}{\textbf{96.1}} & \multirow{2}{*}{\textbf{95.9}} & \multirow{2}{*}{\textbf{96.6}} \\
\hspace{3ex} (\sysname) & & \\
\hline
\end{tabular}
    \caption{{\bf Consistency:} By adding a rational layer to the baseline LLM, \sysname 
    significantly improves
      consistency among beliefs by resolving uncovered conflicts.}  %
\label{Results-overall}
\end{table}

\paragraph{Accuracy.}
Table \ref{Results-acc} shows overall performance on our three datasets
(test partitions). As can be seen, we observe stable accuracy, as well as the answers now being faithful to the
reasoning chains in the belief graph. This is significant, as it
allows users to understand how answers follow from system beliefs (and
in cases where an LLM belief was flipped, why that belief is
untenable in the broader system).

\begin{table}[ht]
\centering
\setlength{\tabcolsep}{2pt}
\begin{tabular}{l|ccc} \hline
   & Entail- &  &  \\
System & mentBank & OBQA & Quartz \\ \hline  
 LLM & 79.4 & 74.0 & 80.2\\ 
 LLM + rational layer & \multirow{2}{*}{79.9} & \multirow{2}{*}{75.0} & \multirow{2}{*}{80.0} \\
 \hspace{3ex} (\sysname)  & & & \\
\hline
\end{tabular}
\caption{{\bf QA accuracy:} \sysname's belief revision in the rational layer 
preserves overall QA accuracy.}
\label{Results-acc}
\end{table}

\paragraph{Ablations.}
  
To study the impact of the three different types of rules on consistency improvement,
we using the EntilmentBank dataset (dev partition). To do this, given the belief graph
for a question, we mask out (separately, rather than cumulatively) each type of rule in turn when providing the graph
to the MaxSAT solver. We then run the
constraint solver and measure the resulting self-consistency of beliefs on the original graph.

\begin{table}[ht]
\centering
\begin{tabular}{lc} \hline
  System & EntailmentBank \\ \hline
\sysname (our system): & 96.1 \\
\hspace*{3mm} - without $p \rightarrow h$ rules & 93.8 \\
\hspace*{3mm} - without XOR rules & 90.4 \\
\hspace*{3mm} - without MC rule & 95.8 \\
\hline
\end{tabular}
    \caption{{\bf Consistency:} Ablations on EntailmentBank (Dev) suggest that all three types of rules contribute to improving self-consistency.
  \label{Results-ablation-type}}
\end{table}

The results are shown in Table~\ref{Results-ablation-type} (the MC rule
is the constraint that exactly one multiple-choice option should be chosen,
Section~\ref{belief-graph-generation}).
The results indicate that all three types of rules contribute
to the system's consistency improvements.

\subsection{Success Analysis \label{successanalysis}}

\begin{figure}[t]
\centering
     \includegraphics[width=1\columnwidth]{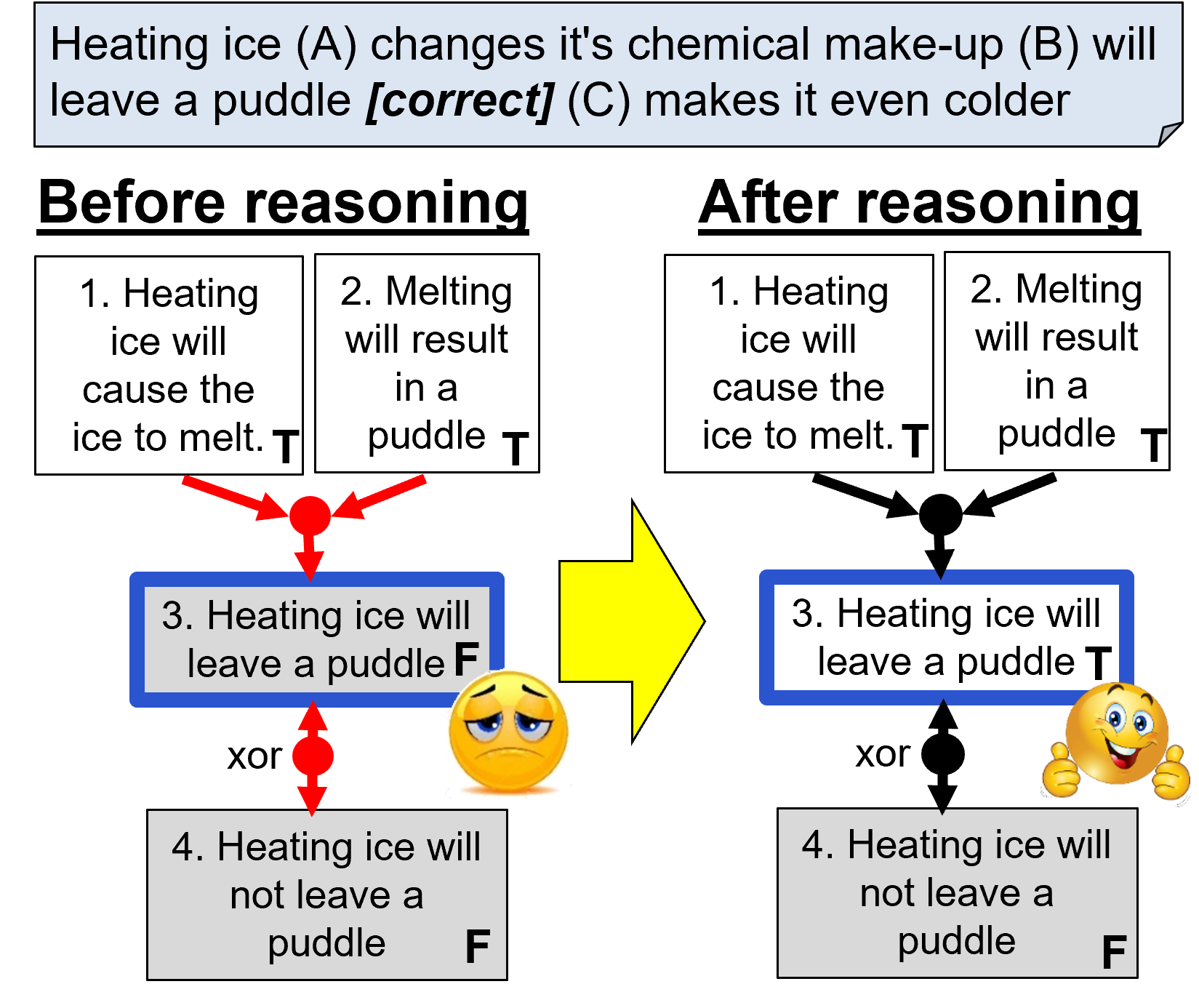}	   %
      \caption{{\bf Example of good reasoning:} The model's beliefs in 1 and 2, and the rule 1 \& 2 $\rightarrow$ 3, as well as the xor constraint, causes it to (desirably) flip its belief in 3 from false (grey, before) to true (white, after). \label{success1}}
\end{figure}

\begin{figure}[t]
\centering
     \includegraphics[width=1\columnwidth]{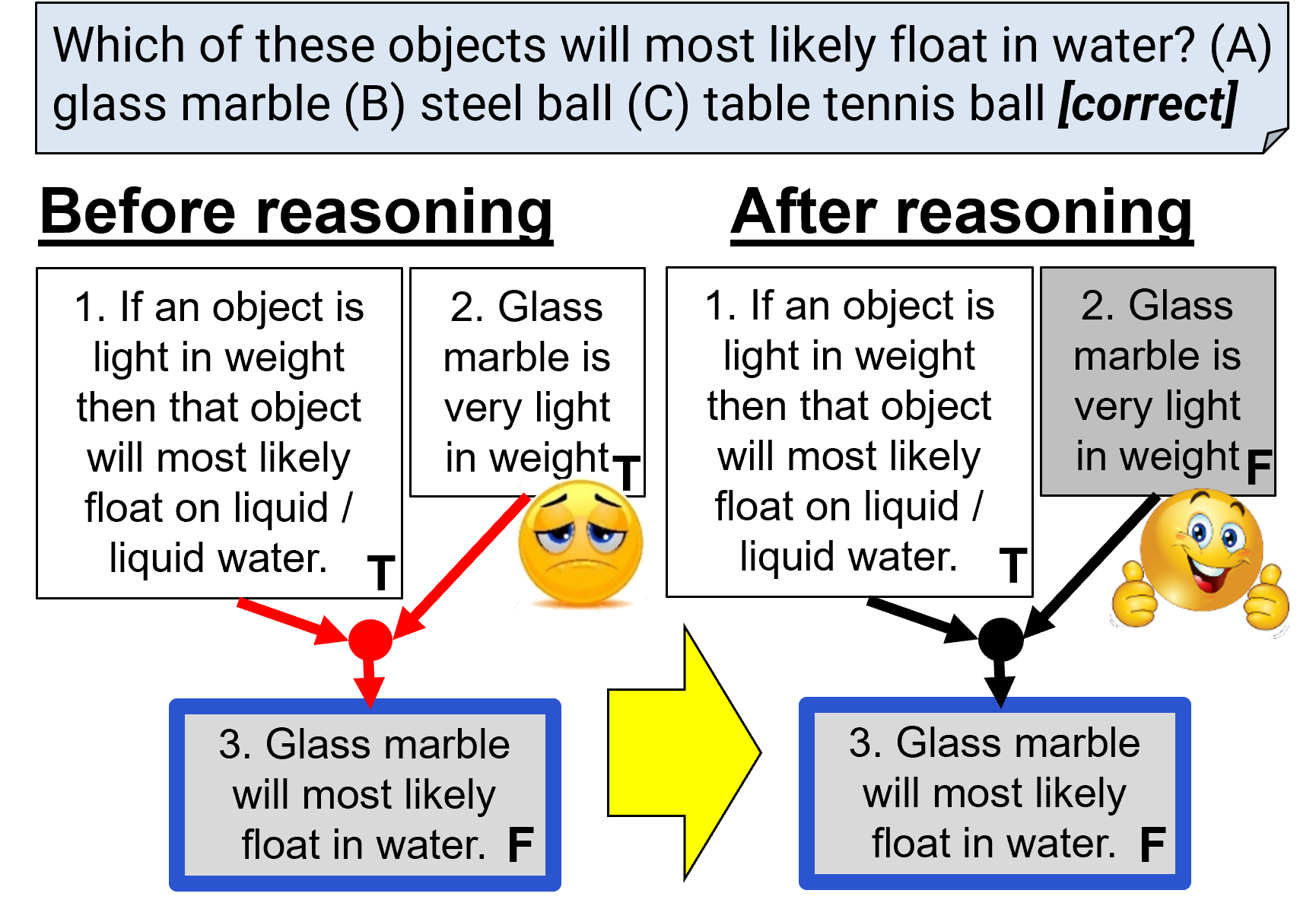}	   %
      \caption{{\bf Example of good reasoning:}
Although the model correctly believes option (A) is false (grey, node 3), this answer conflicts with other beliefs (red). Reasoning leads the system to realize that its weakest belief (2) is actually false, correctly flipping its label from true (white) to false (grey, right side) restoring consistency. \label{success2}}
\end{figure}

\begin{figure}[t]
\centering
     \includegraphics[width=1\columnwidth]{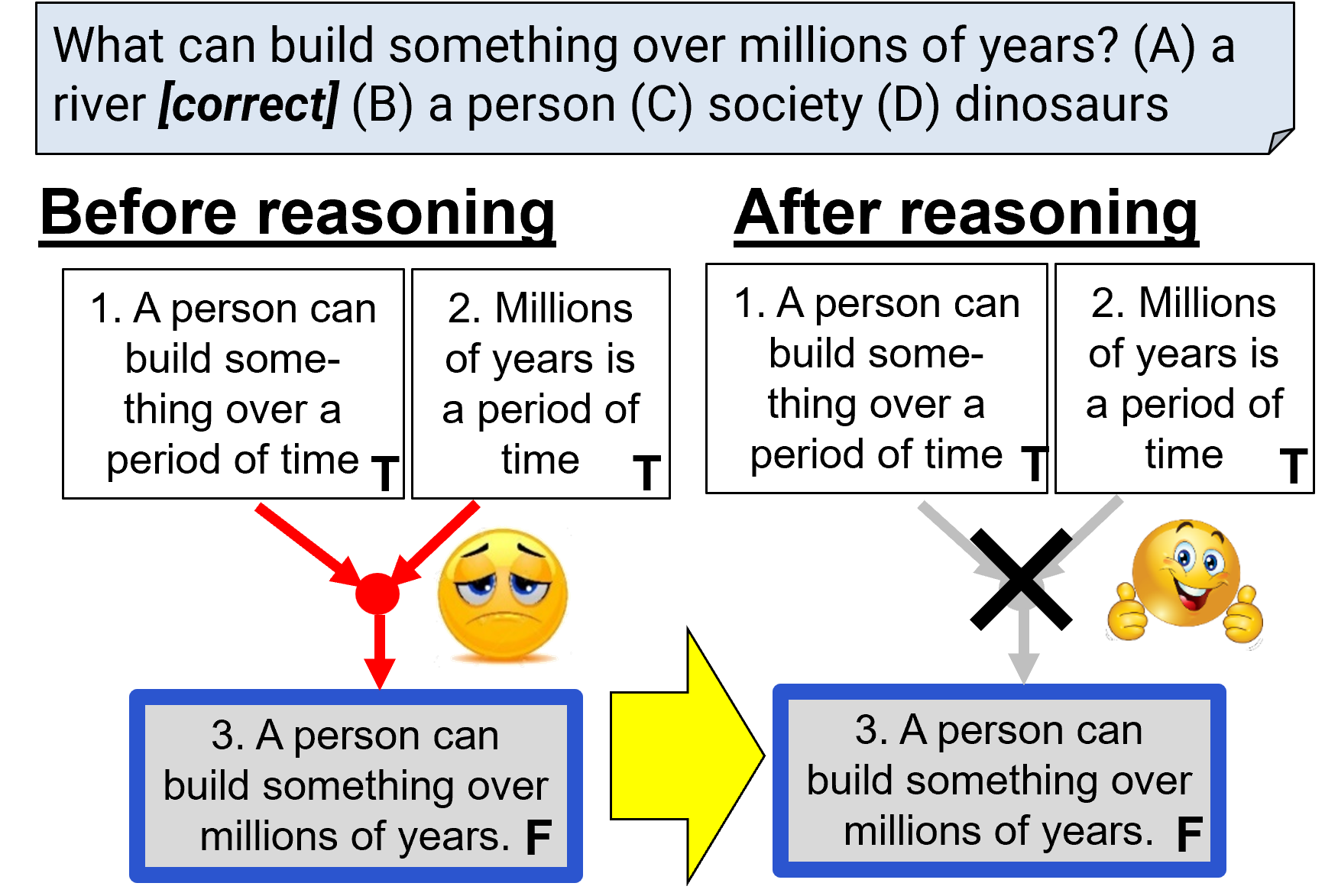}	   %
      \caption{{\bf Example of good reasoning:}
Here the reasoner (desirably) chooses to reject the violated (bad) rule rather than flip a belief, as the minimum cost way to restore consistency. \label{success3}}
\end{figure}

We identify three classes of successful reasoning by the constraint reasoner:
(a) latent model beliefs correct an initially wrong answer (Figure~\ref{success1});
(b) the system corrects an initially erroneous, latent model belief 
(Figure~\ref{success2});
and (c) strong model beliefs identify and reject a bad rule (Figure~\ref{success3}).
These types of system corrections help to improve accuracy
and produce answers supported by valid chains of reasoning, allowing
users insight into why an answer follows from the model's knowledge.

\subsection{Failure Analysis}
\label{erroranalysis}

\begin{figure}[t]
\centering
     \includegraphics[width=1\columnwidth]{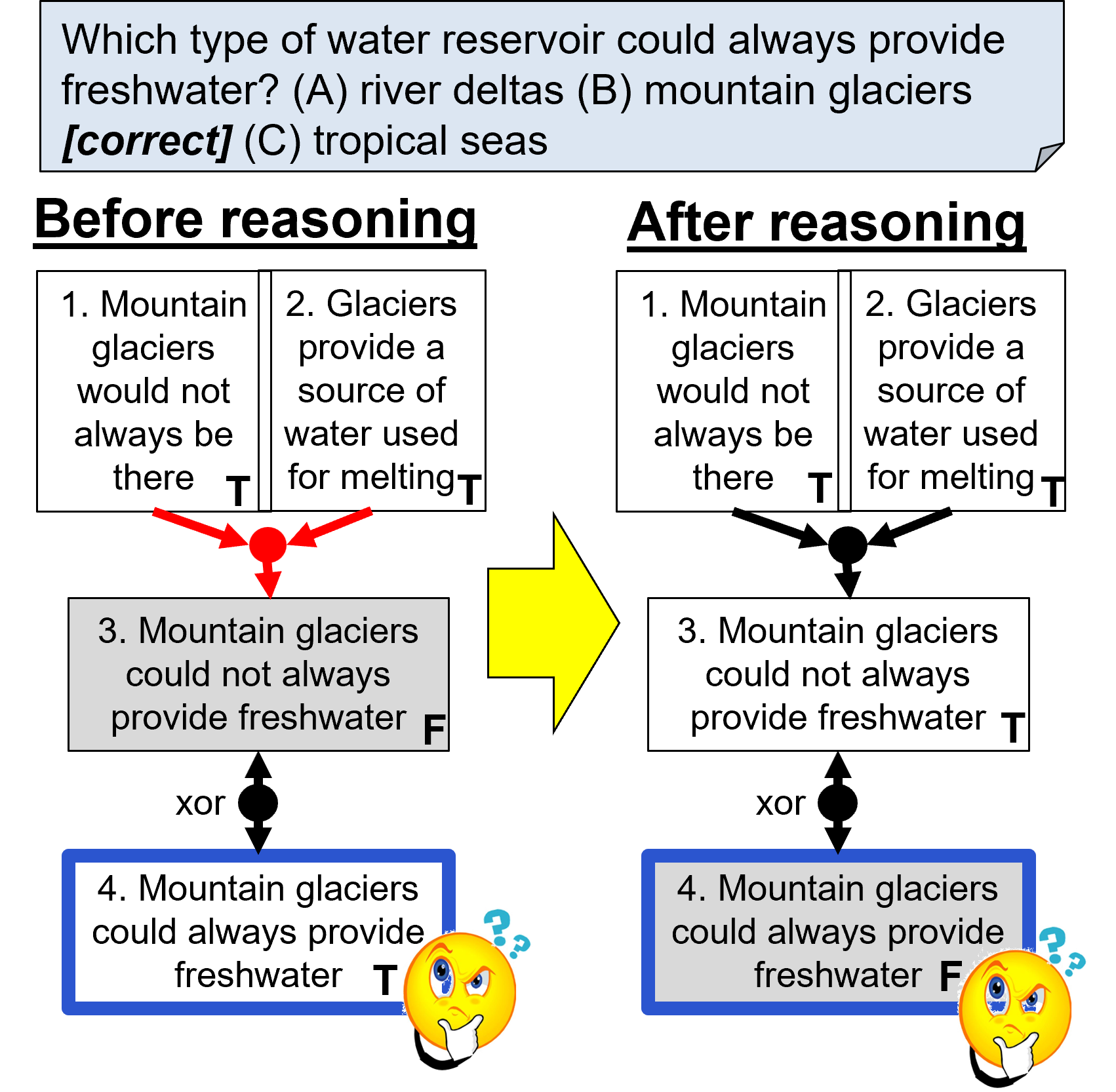}	   %
\caption{{\bf Unexpected reasoning:} Here the model unexpectedly pays particular attention to the
world ``always''. Because it strongly believes that glaciers will not {\it always} be there (1, white),
the system prefers to flip its beliefs in 3 and 4, rather than flipping 1, thus rejecting answer
option B (arguably correctly). \label{failure1}}
\end{figure}

\begin{figure}[t]
\centering
     \includegraphics[width=1\columnwidth]{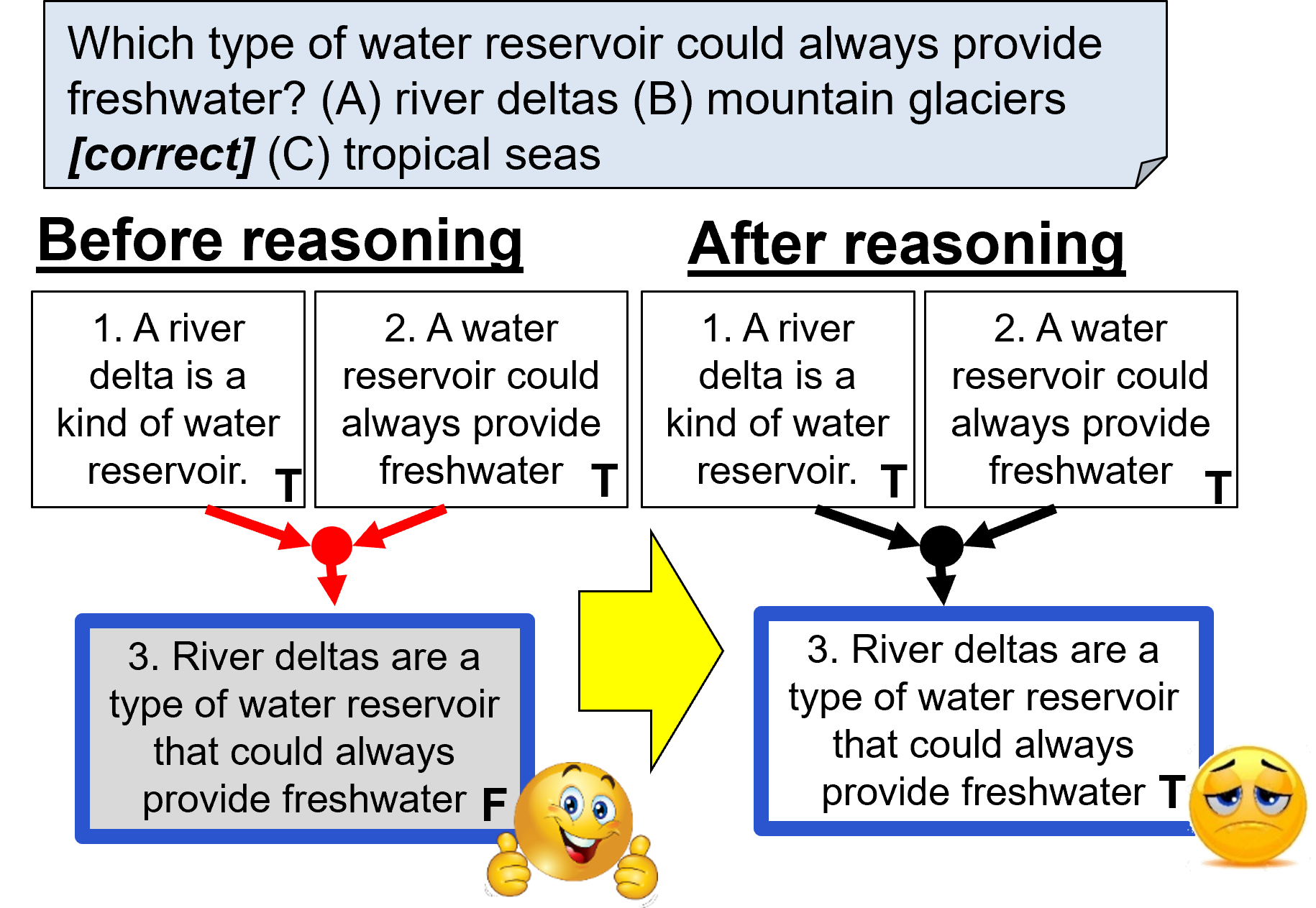}	   %
\caption{{\bf Failure due to bad beliefs:} The model strongly believes both 1 and 2 (although both are factually incorrect), here causing 3's label to undesirably flip from false (correct) to true (incorrect). \label{failure2}}
\end{figure}

Reasoning can also make mistakes. From a manual analysis of 50 random questions from EntailmentBank that \sysname answered incorrectly, we identified five main causes of failure and their approximate frequency
({\bf Note} that multiple categories can apply, hence total is $> 100\%$):

\paragraph{1. Missing Rules ($\approx$30\%):}
In some cases, the system generates irrelevant rules but misses an important
one needed to support the correct answer, resulting
in incorrect conclusions. While somewhat subjective, this is a notable error
category that we observe. For example for the question:
\begin{myquote}
{\it A human cannot survive the loss of %
(A) The liver \bfit{[correct]} (B) A lung (C) A kidney}
\end{myquote}
the system incorrectly concludes (B) is true, ignoring the
commonsense rule that with two lungs, a person can survive without one of them.

\paragraph{2. Incorrect Beliefs ($\approx$30\%):}
Sometimes the reasoner fails to correct incorrect model beliefs,
either because the model's confidence is high or evidence against them is weak or missing.
In the example shown in Figure~\ref{failure2}, the model's strong, incorrect
beliefs that ``river deltas are reservoirs'' and ``reservoirs always provide freshwater''
(untrue of oceans, say) causes it to incorrectly conclude that ``deltas are freshwater reservoirs''.

\paragraph{3. Incorrect Rules ($\approx$10\%):}
Rule generation can produce bad rules, e.g., in Figure~\ref{success3}), and
in some cases the constraint reasoner fails to reject them if they are strongly
believed. In particular, confusion or ambiguity over quantifiers can
result in bad rules, e.g., (emphasis added)
``{\it Some} animals catch their prey with trickery.'' \& 
``A spider is a kind of animal.'' $\rightarrow$
``Spiders catch their prey with trickery.''.
Similarly the model generates the fallacy:
	``Some people don't mind not moving for an hour''  \& 
``breathing is a kind of movement''
$\rightarrow$ ``Some people don't mind not breathing for an hour.''

\paragraph{4. Ambiguous Statements, Unexpected Reasoning ($\approx$10\%):}
A common cause of error is the surprising ambiguity of belief statements, which can
often be read in multiple ways. In several cases, the
model adopts a valid but unexpected interpretation, resulting in ``errors'' compared
to the gold answer label. For example, in Figure~\ref{failure1}, the model
takes the word ``always'' in a literal sense (``glaciers will not {\it always} be there''),
resulting in an answer that differs from the gold label.
Developing ways to attach context to these statements to
help disambiguate them would help alleviate such errors.

\paragraph{5. Multiple Valid Answers ($\approx$10\%):}
A final cause of ``error'' - at least with
respect to the gold label - is that multiple answers may be valid, and the question
is asking for the {\bf best} answer;
eg. for ``What could fill a beach ball? (A) Oxygen (B) Water ...'',
A is labeled correct, while B is also a valid answer. \sysname
(desirably) finds valid reasoning chains for both, but
the notion of highest-scoring proof does not fully correlate
with the notion of ``best answer'' intended by the question author.

\section{Future Work}

There are several impactful ways this work could be
further extended.
First, incorporating the question's \emph{context} in the belief statements in our rational layer could make the semantics of the beliefs more precise, thus avoiding potential ambiguity in their truth value.
Second, one could use the belief graph itself to identify the key reasoning pieces that the LLM is most uncertain about. This could then guide a \emph{human-in-the-loop} mechanism to correct or validate uncertain pieces via user interaction.
Third, maintaining a \emph{persistent belief graph} over multiple questions could help make the system more consistent across questions. This, in turn, would make a user's conversational experience with the system more coherent in a longer dialog setting.
Lastly, after resolving inconsistencies in the rational layer, we could consider \emph{propagating information back to the LLM layer} in order to update it (via fine-tuning, model editing, memory-based architectures, etc.), helping avoid similar inconsistencies in the future.

\section{Conclusion}
While LLMs perform well, the interdependencies between their answers and their other
beliefs is opaque, and may even be in conflict. 
This lack of interpretability is a significant impediment to widespread use of LLMs.
To reduce this opacity, and reduce these conflicts, we have proposed \sysname,
a new system architecture in which an explicit, interpretable representation of
beliefs - the {\bf belief graph} - is added as a {\bf rational layer} above the LLM.
This layer providing a window into system beliefs, and allows
latent inconsistencies in the LLM alone to reasoned about and repaired.
Our implementation shows that belief consistency of the overall system is
significantly improved, without harming answer accuracy,
resulting in answers supported by interpretable chains of reasoning drawn
from a more consistent belief system. 
This new architecture is an important step towards improving confidence in system behavior, and
towards trustable deployment of LLMs in practical applications.

\section*{Limitations}

We have shown how an LLM can be extended with a self-reflective component,
allowing latent model knowledge to be made explicit in the form of a {\bf belief graph},
providing a window into the model’s system of beliefs. While exciting, there
are several limitations with the current work and opportunities for the future.

First, the reasoning component in the rational layer can make mistakes, resulting in the
overall system rejecting true statements or accepting false ones.
A detailed analysis and classification of these failure modes was
presented in Section \ref{erroranalysis}.

Second, for our experiments, we used the T5-11B based Entailer system as the baseline LLM. While there is every reason to expect our proposed architecture to be effective in reducing inconsistency with newer and larger LLMs such as ChatGPT and LLaMA, this is still to be evaluated. Doing so would require implementing the basic operations needed to construct belief graphs (Section~\ref{subsec:basic-ops}) using instruction prompting and in-context learning. Other work has demonstrated such implementations \citep[e.g.,][]{Wei2022ChainOT,alpaqa}, making the outlook promising, but indeed their combination still needs to be demonstrated at scale in an architecture like REFLEX.

Lastly, we found consistency-minimized belief graphs to be highly valuable in understanding the system's successes and failures. We expect these graphs to be a valuable starting point for providing explanations and gaining a user's trust in the system. However, we have not conducted a formal user study to measure this.

\section*{Ethics Statement}

Like any other project using LLMs, despite
the best intentions there is a risk of the model
producing biased or offensive statements as part
of its explanations, and thus must be used with care
and appropriate guards and warnings.

\section*{Acknowledgements}
This research was made possible, in part, by funding from Open Philanthropy, the European Research Council (\#740516) and by the German Federal Ministry of Education and Research (BMBF) under Grant No.\ 01IS18036A.
We also thank Google for providing the TPUs for conducting experiments. 
Finally, we are grateful for the valuable feedback from the anonymous reviewers.

\bibliography{anthology,custom}
\bibliographystyle{acl_natbib}

\clearpage

\appendix

\setcounter{table}{0}
\renewcommand{\thetable}{\thesection\arabic{table}}
\renewcommand{\thefigure}{\thesection\arabic{figure}}

\section{Additional Results}
\label{appendix:addl-results}

We report results on the dev set of the EntailmentBank dataset in Table~\ref{Results-devset}.

\begin{table}[ht]
\centering
\setlength{\tabcolsep}{4pt}
\begin{tabular}{l|cc} \hline
System & \multicolumn{2}{c}{EntailmentBank (dev)} \\ \hline
 & Consistency & Accuracy \\
LLM & 87.5 & 78.6 \\
LLM + rational layer 
 & \multirow{2}{*}{\textbf{96.1}} & \multirow{2}{*}{\textbf{81.8}}\\
 \hspace{3ex} (\sysname) & & \\
\hline
\end{tabular}
\caption{Results on EntailmentBank (dev), used to tune the system's hyperparameters. %
\label{Results-devset}}
\end{table}

\setcounter{table}{0}

\section{Hyperparameters and Runtime}
\label{appendix:hyperparams-and-runtime}

MaxSAT finds the optimal assignment of true/false labels on statement nodes that minimizes the total penalty of constraint violations. If the true/false label on a statement node is flipped, then the penalty is the model confidence $c_{\mathrm{s}}$ in the original label. Similarly if a rule (constraint) is violated by the true/false labels on its associated statements, then the penalty is the model confidence $c_{\mathrm{r}}$ in that rule.

We set a number of hyperparameters to ensure that the various
sources of confidence are appropriately balanced, and tune these on
a development set (EntailmentBank (dev) which is separate from our test sets). We use the same
set of hyperparameters for all test sets.
\begin{enu}
\item[1.] As raw model confidences $c_{\mathrm{s}}$ are highly skewed towards 0 and 1,
          we re-calibrate these with $e^{k.(c_{\mathrm{s}}-1)}$, where k is a fixed hyperparameter. Note, that for the MC and XOR rule, the raw input score $s$ is 1.0.
\item[2.] We calibrate rule confidences in the same way as we calibrate belief confidences but use separate calibration parameters different types of rules namely:
\begin{ite}
\item Entailer rules $p \rightarrow h$
\item XOR rules
\item MC rules
\end{ite}
i.e., the raw rule score $c$ is re-calibrated to confidence $e^{k_{type}.(c-1)}$ where $k_{type}$ is the respective hyperparameter per rule type.
\item[3.] We set three hyperparameters tuning the respective importance of the three different types of rules. Therefore, the final rule score is computed by $c = t_{type}*e^{k_{type}.(c-1)}$ where $t_{type}$ is the respective hyperparameter constant per rule type.
\item[4.]  For xor rules between statements $s_{i}$ and $negs_{i}$,
  we remove (ignore) those where there is significant uncertainty,
  namely where $|score(s_{i}) - score(negs_{i})| \leq m_{xor}$, where $m_{xor}$ is 
  a tuned hyperparmeter.
  \item[5.] Additionally, we tune a damping parameter that downscales rules on the boundary of the graph. Belief nodes involved in these rules are not supported by any premises and should therefore have less influence than rules with strong support.
\item[6.] Finally, we tune the maximum depth $\dmax$ of the belief graph.

\end{enu}

The performance on this dev set partition is shown in Table \ref{Results-devset} and the hyperparameter values are shown in Table \ref{hyper}.

The runtime for MaxSAT constraint solving is fast ($<$1 millisecond per question). However, constructing the belief graph is computationally intensive: Each call to expand or score a node takes $\sim$2 seconds, and our graphs typically contain $\sim$600 nodes, so if these calls were maximally parallelized, with each step growing the graph one level deeper, the runtime would be the maximum graph depth (5) x 2 seconds = $\sim$10 seconds total (or several minutes if a naive sequential implementation were used).

\begin{table}[t!]
\centering
{ %
\begin{tabular}{lccc} \hline
Hyperparameter & Value\\ \hline
$k$ & 9 \\ 
$k_{\mathrm{entailer}}$ & 36 \\ 

$k_{\mathrm{xor}}$ & 30 \\ 
$k_{\mathrm{mc}}$ & 9 \\ 
$t_{\mathrm{entailer}}$ & 1.02  \\ 
$t_{\mathrm{xor}}$ & 1.1 \\ 
$t_{\mathrm{mc}}$ & 0.98 \\ 
$m_{\mathrm{xor}}$ & 0.3 \\ 
$\dmax$ & 5 \\ 
\hline
\end{tabular}
}
\caption{Hyperparameters. \label{hyper}}
\vspace{-3mm}
\end{table}

\setcounter{figure}{0}

\end{document}